\documentclass{article} % For LaTeX2e
\usepackage[preprint]{colm2025_conference}

\usepackage{microtype}
\usepackage{hyperref}
\usepackage{url}
\usepackage{booktabs}
\usepackage{graphicx}
\usepackage{float}

\usepackage{lineno}

\definecolor{darkblue}{rgb}{0, 0, 0.5}
\hypersetup{colorlinks=true, citecolor=darkblue, linkcolor=darkblue, urlcolor=darkblue}

\title{Reasoning Court: Combining Reasoning, Action, and Judgment for Multi-Hop Reasoning}

\author{
Jingtian Wu \quad Claire Cardie \\
Cornell University \\
\texttt{\{jw2349, ctc9\}@cornell.edu}
}

\begin{document}

\ifcolmsubmission
\linenumbers
\fi

\maketitle

\begin{abstract}
While large language models (LLMs) have demonstrated strong capabilities in tasks like question answering and fact verification, they continue to suffer from hallucinations and reasoning errors, especially in multi-hop tasks that require integration of multiple information sources. Current methods address these issues through retrieval-based techniques (grounding reasoning in external evidence), reasoning-based approaches (enhancing coherence via improved prompting), or hybrid strategies combining both elements. One prominent hybrid method, ReAct, has outperformed purely retrieval-based or reasoning-based approaches; however, it lacks internal verification of intermediate reasoning steps, allowing potential errors to propagate through complex reasoning tasks. In this paper, we introduce Reasoning Court (RC), a novel framework that extends iterative reasoning-and-retrieval methods, such as ReAct, with a dedicated LLM judge. Unlike ReAct, RC employs this judge to independently evaluate multiple candidate answers and their associated reasoning generated by separate LLM agents. The judge is asked to select the answer that it considers the most factually grounded and logically coherent based on the presented reasoning and evidence, or synthesizes a new answer using available evidence and its pre-trained knowledge if all candidates are inadequate, flawed, or invalid.  Evaluations on multi-hop benchmarks (HotpotQA, MuSiQue) and fact-verification (FEVER) demonstrate that RC consistently outperforms state-of-the-art few-shot prompting methods without task-specific fine-tuning.
\end{abstract}

\section{Introduction}

\begin{figure*}
    \centering
    \includegraphics[width=1\linewidth]{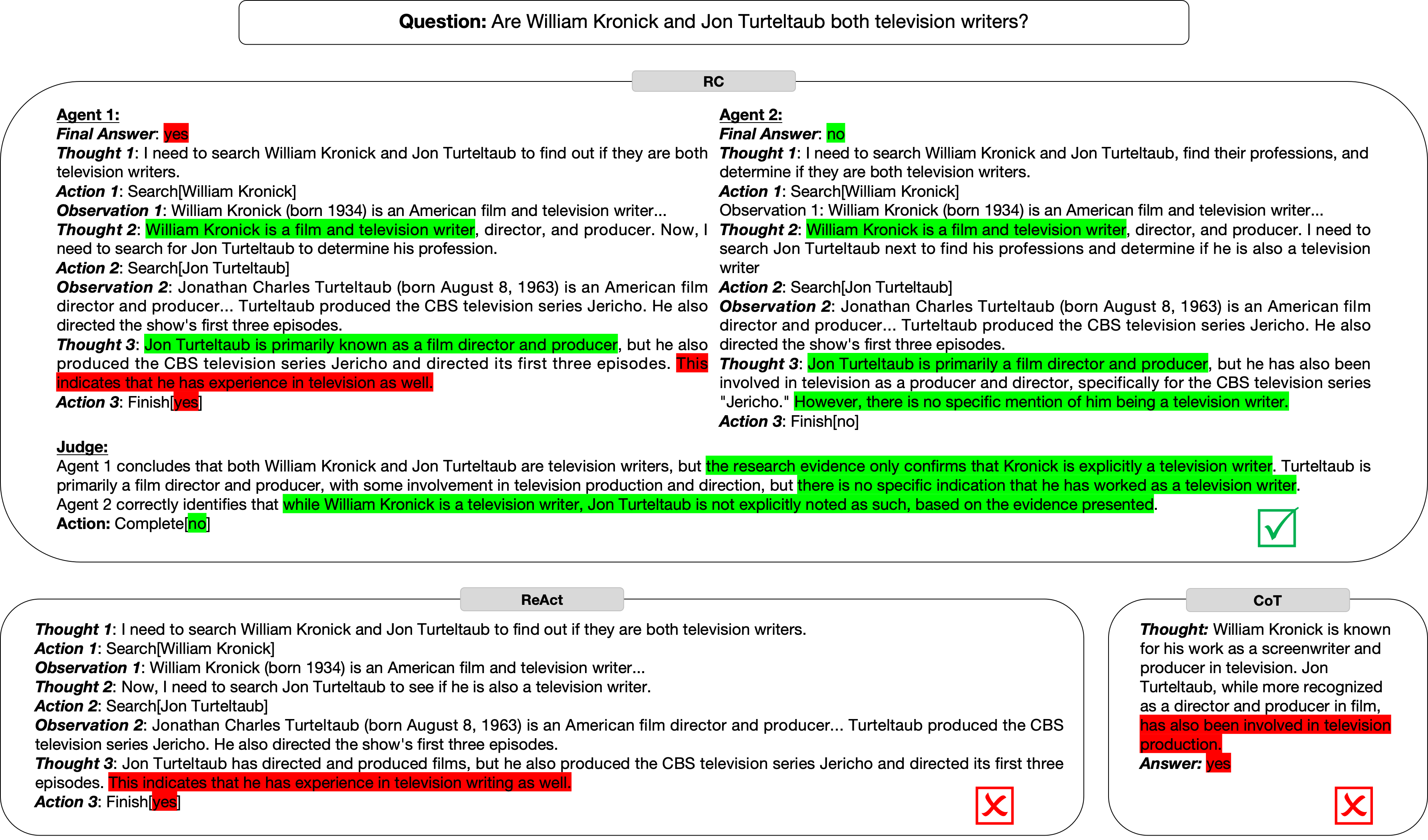}
    \caption{Comparison of RC, ReAct, and CoT methods in answering a HotpotQA \citep{yang2018hotpotqadatasetdiverseexplainable} question. The reasoning and acting stages are labeled as "Thoughts" and "Actions," respectively. Evidence, containing information retrieved from Wikipedia, is presented in "Observations." The final answer provided by the agent is shown in "Final Answer." Red highlights indicate incorrect reasoning or decisions made by the LLM agent, whereas green highlights represent correct reasoning or decisions.}
    \label{fig:qa-comparison}
\end{figure*}

Large language models (LLMs) have demonstrated significant improvements in multi-step reasoning and problem-solving, enabling them to handle complex question-answering tasks with increased accuracy \citep{aksitov2023restmeetsreactselfimprovement, smit2024goingmadlookmultiagent, yao2023react}. However, despite these advancements, LLMs continue to face challenges in multi-hop reasoning, where integrating information from multiple sources and reasoning steps is crucial for reaching accurate conclusions \citep{lee2022generativemultihopretrieval, yao2023react}. These challenges often manifest as hallucinations, where models generate false or fabricated information, and reasoning errors, where models fail to coherently integrate and interpret retrieved evidence \citep{yao2023react}, as illustrated in ReAct and CoT examples in Figure~\ref{fig:qa-comparison}.

Earlier methods, such as chain-of-thought (CoT) prompting \citep{wei2023chainofthoughtpromptingelicitsreasoning}, rely solely on internal reasoning without information retrieval. Consequently, as illustrated in Figure~\ref{fig:qa-comparison}, CoT cannot reliably determine whether the individual explicitly works as a television writer, instead erroneously equating general television involvement with writing. Building upon this purely reasoning-based approach, recent few-shot prompting methods have tackled multi-hop reasoning challenges by combining retrieval with reasoning. For example, ReAct \citep{yao2023react} interleaves reasoning steps and actions to ground outputs in external evidence; however, as shown in Figure~\ref{fig:qa-comparison}, it still produces reasoning errors, mistakenly concluding that both individuals are television writers despite accurately retrieving their career details.

To address these limitations, we introduce Reasoning Court (RC), a framework developed in a few-shot prompt-based setting. RC draws inspiration from ReAct \citep{yao2023react} for its integration of retrieval and reasoning, and the evaluative paradigm of LLM-as-a-Judge \citep{zheng2023judgingllmasajudgemtbenchchatbot}. RC employs an LLM judge to systematically assess the intermediate reasoning steps produced by multiple agents, each of which iteratively interleaves reasoning with evidence retrieval. The judge is asked to evaluate these research trajectories—structured sequences of reasoning steps, retrieval actions, and retrieved evidence (observations)—to select the answer it considers to be the most factually supported and logically coherent. A distinctive strength of RC’s judge is its ability to move beyond simple answer selection: when it considers all candidate answers to be inadequate or flawed, it independently synthesizes a response by leveraging retrieved external evidence and, if necessary, its pre-trained knowledge. As illustrated in Figure~\ref{fig:qa-comparison}, RC first uses two agents to generate candidate answers from retrieved evidence. In this example, Agent 1 incorrectly concludes "yes" by inferring that Turteltaub’s television production roles imply that he is a writer, while Agent 2 correctly concludes "no" by noting that only Kronick is explicitly a television writer. In the judgment phase, the LLM judge identifies Agent 1’s flawed reasoning and selects Agent 2’s answer, resulting in the correct final response.

Our empirical evaluation across benchmarks like HotpotQA \citep{yang2018hotpotqadatasetdiverseexplainable}, FEVER \citep{thorne2018feverlargescaledatasetfact}, and MuSiQue \citep{trivedi2022musiquemultihopquestionssinglehop} demonstrates RC's effectiveness in multi-hop reasoning tasks. RC consistently outperforms the strongest few-shot prompting baselines, achieving significant improvements in exact match (EM) and F1 scores.

\section{Related Work}

\paragraph{Language Models for Debate} Debate mechanisms have been widely studied as a means to improve reasoning quality and factual accuracy in large language models (LLMs). \citet{du2023improvingfactualityreasoninglanguage} introduced a multi-agent debate approach, where multiple LLMs iteratively generate, critique, and refine their responses to enhance performance in reasoning-intensive tasks such as mathematical problem-solving and strategic decision-making. Building on this research, \citet{liang2024encouragingdivergentthinkinglarge} introduced the Multi-Agent Debate (MAD) framework to address the Degeneration-of-Thought (DoT) problem by encouraging divergent thinking through debates between LLMs, with each model presenting and challenging arguments under the supervision of a judging model. In addition, \citet{khan2024debatingpersuasivellmsleads} explored the effectiveness of debates between more persuasive LLMs in producing more truthful answers.

However, some studies suggest that debate mechanisms may not always be beneficial in practice. \citet{parrish2022singleturndebatedoeshelp} demonstrated that single-turn debate explanations, where both correct and incorrect answers are argued for, do not improve human performance on challenging reading comprehension tasks. Further exploring this, \citet{parrish2022twoturndebatedoesnthelp} found that even a two-turn debate, which includes counter-arguments, does not significantly enhance human decision-making accuracy. These findings raise concerns about the potential limitations of debate-style mechanisms, especially when employed in LLM systems intended to assist humans in reasoning tasks. Additionally, \citet{smit2024goingmadlookmultiagent} evaluated multiple multi-agent debate (MAD) configurations across diverse datasets and found that, although debate-based approaches can be enhanced to achieve competitive performance, their success is highly dependent on hyperparameter tuning. Furthermore, the inherent complexity of the debate process not only increases computational costs, but also introduces additional dynamics that can skew the final judgment.

\textit{Unlike debate-driven reasoning, RC does not rely on adversarial interactions between LLMs. RC directly employs an LLM judge to evaluate research trajectories generated by LLM agents.} This streamlined approach omits the multi-agent debate process, reducing computational overhead and mitigating the risk of persuasive yet misleading arguments that could compromise the judge’s impartiality.

\paragraph{Language Models for Retrieval} Retrieval-based mechanisms (including the combination of reasoning and retrieval) have been widely explored to reduce hallucinations and improve the factual accuracy of LLM-generated responses. \citet{lee2022generativemultihopretrieval} introduce Generative Multi-hop Retrieval (GMR) to generate retrieval sequences within the model’s parametric space. \citet{yao2023react} introduced the ReAct paradigm, which combines reasoning and acting in LLMs by interleaving reasoning traces and task-specific actions to enhance interaction with external environments and improve performance on various tasks. Building on ReAct, \citet{aksitov2023restmeetsreactselfimprovement} presented a ReAct-style LLM agent that integrates a self-improvement framework through Reinforced Self-Training (ReST) to refine the agent’s reasoning and actions iteratively. \textit{RC builds on and extends ReAct by introducing an additional judge component to evaluate its reasoning process and either select or synthesize an answer that it considers more logically coherent based on the retrieved evidence.}

\paragraph{Language Models for Evaluation and Judging} Studies have examined the role of LLMs as evaluators or judges in multi-turn conversations. \citet{zheng2023judgingllmasajudgemtbenchchatbot} introduced the LLM-as-a-judge framework, demonstrating that models like GPT-4 can effectively act as judges, achieving over 80\% agreement with human preferences. This method provides a scalable and explainable alternative to human evaluations, and aligns with the goal of ensuring that LLM-based judgments are consistent and grounded in quality assessments. \textit{Unlike typical judge-based frameworks that solely select from existing candidate responses, the judge in RC can synthesize a new answer based on retrieved evidence and its pre-trained knowledge when it determines that none of the existing answers is sufficiently coherent or well-supported by evidence.}

\section{Method}
This section presents the two core phases of the RC framework: (1) the reasoning and retrieval phase, and (2) the judgment phase.

\subsection{Reasoning and Retrieval Phase}
The reasoning and retrieval phase of RC is based on the ReAct framework \citep{yao2023react}, which alternates between reasoning and retrieval to solve complex tasks. 
At each time step $t$, the model receives an observation $o_t \in O$ and takes an action $a_t \in A$ based on a policy $\pi(a_t | c_t)$, where the context $c_t = (o_1, a_1, \dots, o_{t-1}, a_{t-1}, o_t)$. The reasoning and retrieval phase proceeds for up to 7 steps for HotpotQA and MuSiQue, and up to 5 steps for FEVER, consistent with the limits set by \citet{yao2023react}, who empirically showed that additional steps beyond these limits do not further improve performance. In particular, given a query \( q \), RC employs two agents that independently generate answers \( a_1 \) and \( a_2 \) through interleaved reasoning and retrieval. To enhance efficiency and scalability, the agents run concurrently.

RC adopts ReAct's action space, with slight modifications for the MuSiQue dataset. For HotpotQA and FEVER, the action space includes three types of actions \citep{yao2023react}: (1) \textit{Search[entity]}, which retrieves the first five sentences from a Wikipedia page matching the specified entity, or alternatively suggests up to five most related entities if an exact match is unavailable; (2) \textit{Lookup[string]}, which returns the next occurrence of a sentence containing the specified string; and (3) \textit{Finish[answer]}, which finishes the task with \textit{answer}. For MuSiQue, the action space includes: (1) \textit{Lookup[title]}, which retrieves the content of a paragraph based on its title; and (2) \textit{Finish[answer]}, which concludes the task with \textit{answer}.

\subsection{Judgment Phase}
\paragraph{Input}
The judge receives a query \( q \), the final answers \( a_1 \) and \( a_2 \) provided by the agents, and
their corresponding research trajectories \( \tau_1 \) and \( \tau_2 \).

\paragraph{Evaluation}
The judge is asked to evaluate each solution by verifying that its reasoning is based solely on the provided evidence—free from logical errors, unsupported assumptions, or hallucinations—and that its conclusions are justified by that evidence. If both fail to effectively address the question, the judge derives a specific final answer from the evidence (or, if necessary, its pre-trained knowledge). When the answers differ, the one based on more accurate and coherent reasoning is selected, with a brief explanation of the choice. The final answer is output in the format \textit{Complete[\textless answer\textgreater]}. For the exact prompts, see Appendix~\ref{rc_judge}.

\section{Experimental Setup}

This section details the experimental design used to evaluate RC, including the datasets, baselines, and model configurations.

\subsection{Datasets}
We evaluate Reasoning Court (RC) on three challenging multi-hop reasoning benchmarks: FEVER \citep{thorne2018feverlargescaledatasetfact}, HotpotQA \citep{yang2018hotpotqadatasetdiverseexplainable}, and MuSiQue \citep{trivedi2022musiquemultihopquestionssinglehop}. These benchmarks are chosen to evaluate RC across increasing levels of difficulty. FEVER tests fact verification and grounding answers in single pieces of evidence. HotpotQA, evaluated in the fullwiki setting, requires retrieving evidence from the entire Wikipedia. MuSiQue, composed of questions requiring multiple reasoning hops across 20 paragraphs with mixed relevant passages and distractors, tests RC's ability to query paragraph titles and integrate retrieved content effectively.

For all datasets, we randomly sample a subset of 500 validation questions for evaluation with GPT-4o-mini, following the same selection process used in ReAct \citep{yao2023react}, ensuring reproducibility with a consistent random seed. For Claude, we evaluate on the first 100 questions from this set due to budget constraints.

\subsection{Baselines}
We evaluate RC against several general-purpose LLM baselines in a few-shot prompt-based setting using Exact Match (EM) and F1 scores. The baselines include:
(1) \textbf{Standard Prompting}: A basic prompting approach without structured reasoning or retrieval integration.
(2) \textbf{Chain-of-Thought (CoT)} \citep{wei2023chainofthoughtpromptingelicitsreasoning}: A reasoning-based approach that structures the reasoning process through sequential prompts.
(3) \textbf{Chain-of-Thought with Self-Consistency (CoT-SC)} \citep{wang2023selfconsistencyimproveschainthought}: An extension of CoT that enhances reasoning through self-consistency.
(4) \textbf{MAD} \citep{liang2024encouragingdivergentthinkinglarge}: A method where two agents debate iteratively without retrieval, and a judge oversees the process to select the final answer based on their arguments.
(5) \textbf{ReAct} \citep{yao2023react}: A method that combines reasoning with actions to dynamically plan, retrieve, and update information.
(6) \textbf{Hybrid Approaches}: Combinations like ReAct $\to$ CoT-SC \citep{yao2023react}, CoT-SC $\to$ ReAct \citep{yao2023react}, and ReAct $\to$ Self-Refine that combine ReAct with reasoning-based techniques sequentially.

\subsection{Model Configurations}
For all experiments, we use the GPT-4o-mini and Claude-3.5-Sonnet-20241022 models as the underlying language models. Although we explored the open-source model Llama, we excluded it from our experiments due to its inability to align with the ReAct framework's few-shot prompting methodology (e.g., it failed to produce coherent research trajectories; see Appendix~\ref{exclusion} for details). To ensure fairness across frameworks, we adopt consistent configurations. For CoT-SC methods, we use 21 self-consistency samples with a temperature of 0.7 \citep{yao2023react}; for other methods, the temperature is set to 0.

For ReAct $\to$ CoT-SC, if ReAct fails to return an answer within a predetermined number of steps (7 for HotpotQA and MuSiQue, 5 for FEVER), the system transitions to CoT-SC. For CoT-SC $\to$ ReAct, if the majority answer from the self-consistency samples appears less than 50\% of the time, the system switches to ReAct \citep{yao2023react}. In ReAct $\to$ Self-Refine, the model first generates an answer using ReAct, then iteratively refines it with self-feedback \citep{madaan2023selfrefineiterativerefinementselffeedback}.

\section{Results}
This section presents the empirical evaluation of RC, including its performance on benchmark datasets, the effectiveness of the RC judge, and an analysis of overall computational efficiency.

\subsection{Evaluation on Benchmark Datasets}

Table~\ref{tab:combined_hotpotqa_fever_musique} summarizes RC’s performance on HotpotQA, FEVER, and MuSiQue compared to the baseline approaches. Overall, RC consistently outperforms the strongest few-shot prompting baselines.

On HotpotQA, RC achieves notably higher Exact Match (EM) and F1 scores compared to methods such as ReAct $\to$ CoT-SC and CoT-SC, demonstrating its superior integration of retrieval and reasoning. Similarly, on FEVER, RC’s robust fact-checking capabilities result in significantly higher EM scores than those of ReAct and CoT-based approaches. On MuSiQue, RC again leads in both EM and F1, underscoring its effectiveness in handling complex multi-hop reasoning tasks.

\begin{table}
\centering
\begin{tabular}{lccccc}
\toprule
 & \multicolumn{2}{c}{HotpotQA} & FEVER & \multicolumn{2}{c}{MuSiQue} \\ 
 & \textbf{EM (\%)} & \textbf{F1 (\%)} & \textbf{EM (\%)} & \textbf{EM (\%)} & \textbf{F1 (\%)} \\ 
\midrule
Standard Prompting & 28.4 / 34.0 & 41.8 / 47.5 & 60.4 / 62.2 & 3.8 / 10.0 & 15.3 / 19.3 \\ 
CoT & 34.4 / 36.0 & 48.8 / 44.4 & 63.0 / 69.6 & 8.6 / 13.0 & 18.6 / 15.4 \\ 
CoT-SC & 38.0 / 44.0 & 52.9 / 54.5 & 64.0 / 69.2 & 10.6 / 13.0 & 23.1 / 15.3 \\ 
ReAct & 36.2 / 37.0 & 48.7 / 43.4 & 64.8 / 47.0 & 30.4 / 33.0 & 40.6 / 42.0 \\ 
MAD & 34.0 / - & 49.3 / - & 59.4 / - & 7.6 / - & 18.2 / - \\ 
ReAct $\to$ CoT-SC & 40.6 / 44.0 & 56.1 / 46.8 & 65.4 / 54.0 & 34.0 / 37.0 & 45.9 / 44.9 \\ 
CoT-SC $\to$ ReAct & 38.2 / 42.0 & 51.5 / 48.6 & 63.6 / 50.0 & 27.6 / 37.0 & 39.0 / 44.1 \\ 
ReAct $\to$ Self-Refine & 35.8 / 42.0 & 48.0 / 51.2 & 66.8 / 69.0 & 30.6 / 32.0 & 44.5 / 43.3 \\ 
\midrule
RC & \textbf{42.2} / \textbf{48.0} & \textbf{57.1} / \textbf{59.5} & \textbf{74.0} / \textbf{73.0} & \textbf{36.0} / \textbf{42.0} & \textbf{50.0} / \textbf{55.4} \\ 
\bottomrule
\end{tabular}
\caption{Performance comparison on HotpotQA, FEVER, and MuSiQue datasets across GPT-4o-mini and Claude-3.5-Sonnet-20241022 models. Results for each cell are presented in the format \textbf{GPT-4o-mini / Claude-3.5-Sonnet-20241022}, where the value to the left of `/' corresponds to the mean performance of three runs for GPT-4o-mini and the value to the right corresponds to the single run for Claude due to budget constraints. MAD was not evaluated for Claude due to high cost and poor performance, primarily stemming from a lack of retrieval. }
\label{tab:combined_hotpotqa_fever_musique}
\end{table}

\subsection{Judge Evaluation}
\label{judge_eval}
Beyond aggregate performance, we further analyze the role of the RC judge in resolving conflicting outputs from the reasoning agents. The performance of the judge is evaluated on the subset of questions where the two agents provide non-identical or empty answers.

As shown in Table~\ref{tab:judge_eval}, the judge consistently outperforms Standard Prompting and ReAct across all datasets, achieving substantial improvements in both EM and F1 scores. In particular, the improvements over Standard Prompting underscore the value of the agent-generated trajectories, which provide rich context for decision making. Moreover, the judge’s superior performance relative to ReAct highlights the benefit of evaluating research trajectories from both agents, allowing it to make a decision it considers more logically coherent and factually accurate.

\begin{table*}[t]
\centering
\begin{tabular}{lccccc}
\toprule
 & \multicolumn{2}{c}{HotpotQA} & FEVER & \multicolumn{2}{c}{MuSiQue} \\ 
 & \textbf{EM (\%)} & \textbf{F1 (\%)} & \textbf{EM (\%)} & \textbf{EM (\%)} & \textbf{F1 (\%)} \\ \midrule
Standard Prompting & 18.5 & 31.6 & 56.2 & 3.9 & 12.2 \\
ReAct & 22.0 & 29.5 & 53.8 & 20.8 & 28.9 \\
\midrule
Judge & \textbf{28.2} & \textbf{42.7} & \textbf{66.1} & \textbf{26.0} & \textbf{39.4} \\
\bottomrule
\end{tabular}
\caption{Evaluation of the Judge's accuracy on HotpotQA, FEVER, and MuSiQue datasets using GPT-4o-mini. The results are based exclusively on questions where the two agents in RC provided non-identical or empty answers. The table compares the Judge's performance in these challenging scenarios against Standard Prompting and ReAct.}
\label{tab:judge_eval}
\end{table*}

Further analysis on FEVER and HotpotQA (see Table~\ref{tab:judge_error_analysis}) shows that when one agent is correct and the other is incorrect, the judge selects the correct answer with high accuracy (84.2\% in FEVER, 90.6\% in HotpotQA). Even when both agents fail, it deduces the correct answer in 14.7\% and 7.0\% of cases, respectively—a feat impossible for baselines like ReAct or CoT. Illustrative examples are given in Appendix~\ref{error_analysis}. However, when both agents provide the same incorrect answer, the judge rarely corrects them (1/88 in FEVER, 0/115 in HotpotQA). This reveals its strength in resolving disagreements, but also its weakness in overturning incorrect consensus.

\begin{table*}[h]
\centering
\begin{tabular}{lcccc}
\toprule
\textbf{Dataset} & \textbf{Scenario} & \textbf{Total Cases} & \textbf{Correct Judgments} & \textbf{Accuracy (\%)} \\ \midrule
FEVER    & First Scenario              & 95  & 80  & 84.2\% \\
         & Second Scenario & 34  & 5   & 14.7\% \\
         & Third Scenario         & 88  & 1   & 1.1\%  \\ \midrule
HotpotQA & First Scenario              & 53  & 48  & 90.6\% \\
         & Second Scenario & 143 & 10  & 7.0\%  \\
         & Third Scenario         & 115 & 0   & 0.0\%  \\
\bottomrule
\end{tabular}
\caption{Error analysis of the judge on the FEVER and HotpotQA datasets. The first scenario, \textit{One correct, one incorrect}, includes cases where one agent provides a correct answer while the other does not. The second scenario, \textit{Different incorrect answer or both empty}, includes cases where both agents provide incorrect answers that differ or where both answers are empty. The third scenario, \textit{Same non-empty incorrect answer}, occurs when both agents provide the same non-empty incorrect answer.}
\label{tab:judge_error_analysis}
\end{table*}

\subsection{Efficiency Analysis}
\label{subsec:rc_analysis}
Unlike ReAct $\to$ CoT-SC, where ReAct falls back to CoT-SC when it returns an empty answer, the RC judge can synthesize an answer even when both agents return empty responses. This design reduces the overall usage of LLM. For example, as shown in Table~\ref{tab:efficiency_comparison}, on HotpotQA with GPT-4o-mini, RC averages 8.8 LLM calls per question compared to 9.81 for ReAct $\to$ CoT-SC, while maintaining superior accuracy with only a marginal increase in processing time (10.58s vs. 9.53s). This comparison demonstrates that RC reduces LLM usage costs without incurring significant latency, making it both reliable and cost-effective for real-world applications.

\begin{table*}[t]
\centering
\begin{tabular}{lcc}
\toprule
Method & Avg. Time per Question (s) & Avg. LLM Calls per Question \\ \midrule
ReAct $\to$ CoT-SC & 9.53 & 9.81 \\
RC & 10.58 & 8.8 \\ \bottomrule
\end{tabular}
\caption{Efficiency comparison between RC and ReAct $\to$ CoT-SC on HotpotQA using GPT-4o-mini.}
\label{tab:efficiency_comparison}
\end{table*}

\section{Ablation Study}

\begin{figure*}
    \centering
    \includegraphics[width=1\linewidth]{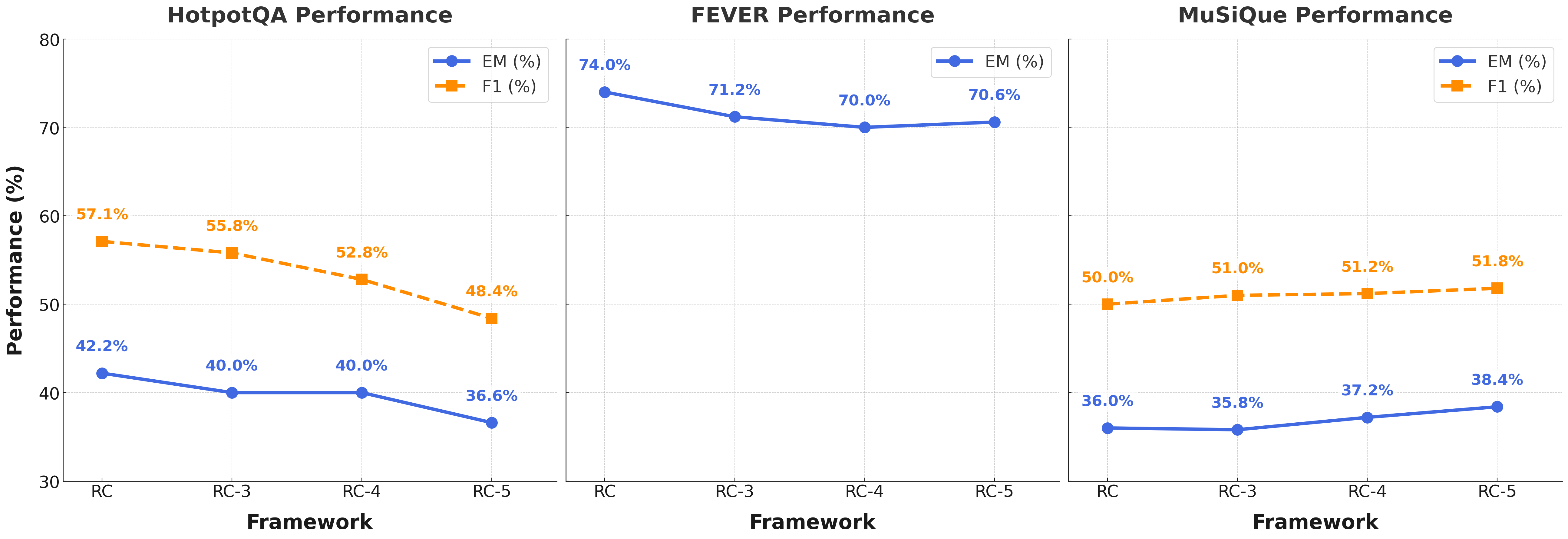}
    \caption{Impact of increasing the number of agents on EM and F1 scores across HotpotQA, FEVER, and MuSiQue. RC represents two agents with LLM temperature set to 0, while RC-3, RC-4, and RC-5 represent 3, 4, and 5 agents respectively, using an LLM temperature of 0.7 to induce diversity in reasoning paths.}
    \label{fig:diversity}
\end{figure*}

This ablation study evaluates the contribution of key components within the Reasoning Court (RC) framework by systematically removing, altering, or adding elements to assess their individual impact. The results, presented in Table~\ref{tab:ablation_study} and Figure~\ref{fig:diversity}, show how each modification affects performance across the HotpotQA, FEVER, and MuSiQue benchmarks.

\begin{table*}[t]
\centering
\begin{tabular}{lccccc}
\toprule
 & \multicolumn{2}{c}{HotpotQA} & FEVER & \multicolumn{2}{c}{MuSiQue} \\ 
 & \textbf{EM (\%)} & \textbf{F1 (\%)} & \textbf{EM (\%)} & \textbf{EM (\%)} & \textbf{F1 (\%)} \\ \midrule
RC (without judge) & 36.2 & 48.7 & 70.0 & 30.4 & 40.6 \\
ReAct-SC & 38.6 & 52.0 & 65.8 & 30.6 & 41.0 \\
ReAct $\to$ MAD & 39.8 & 54.1 & 68.8 & 32.8 & 45.3 \\
ReAct $\to$ LLM-as-a-Judge & 37.4 & 52.1 & 66.6 & 31.0 & 42.1 \\
CoT $\to$ judge & 36.4 & 47.9 & 64.8 & 10.2 & 21.2 \\ \midrule
RC & \textbf{42.2} & \textbf{57.1} & \textbf{74.0} & \textbf{36.0} & \textbf{50.0} \\
\bottomrule
\end{tabular}
\caption{Ablation study results on HotpotQA, FEVER, and MuSiQue datasets using GPT-4o-mini.}
\label{tab:ablation_study}
\end{table*}

\subsection{Impact of the Judge}

The judge is a critical component of RC, tasked with evaluating the research trajectories. When the judge is omitted (\textit{RC without judge}), performance drops significantly across all benchmarks. This finding underscores the judge's essential role in evaluating whether the final answer is both logically consistent and factually accurate. Without the judge, reasoning errors are more likely to directly lead to an incorrect final answer, ultimately reducing overall performance.

\subsection{Comparison Between RC and ReAct-SC}

ReAct-SC uses three agents that work independently without employing a judge. Self-consistency is applied to select the most consistent answer. Our results show that RC outperforms ReAct-SC across all benchmarks, with absolute EM improvements of +3.6\% on HotpotQA, +8.2\% on FEVER, and +5.4\% on MuSiQue. This demonstrates that a structured evaluation by a judge leads to more reliable outcomes than simply relying on self-consistency.

\subsection{Comparison Between RC and ReAct $\to$ MAD}

Compared to RC, ReAct $\to$ MAD adds a debate phase before the final judgment. In this setup, agents argue for their answers citing evidence from their trajectories before the judge selects the final answer (see Appendix~\ref{debate_prompt} for the prompts given to the agents involved in MAD). However, RC consistently outperforms ReAct $\to$ MAD on all benchmarks while being more cost-efficient, as the added debate phase in ReAct $\to$ MAD increases LLM calls. The debate mechanism underperforms because debaters cannot provide new evidence beyond what they have already retrieved and can only introduce additional noise that might hinder the judge's decision. This result aligns with findings from previous studies, such as Smit et al. \cite{smit2024goingmadlookmultiagent} and Parrish et al. \cite{parrish2022singleturndebatedoeshelp, parrish2022twoturndebatedoesnthelp}, which question the effectiveness of debate mechanisms in LLM frameworks. Our findings further reinforce that a well-implemented judge can resolve reasoning discrepancies effectively without requiring a debate phase.

\subsection{Comparison Between RC and ReAct $\to$ LLM-as-a-Judge}

The ReAct $\to$ LLM-as-a-Judge framework employs a standard judge that selects the best answer solely based on relevance and accuracy by considering only the final output of the agents. In contrast, the RC judge evaluates the entire research trajectory by scrutinizing the logical coherence and factual grounding of the agents’ reasoning paths. This evaluation not only aims to uncover potential inconsistencies and errors that might be overlooked when focusing solely on the final answers, but also enables the judge to synthesize a new answer when it believes that neither agent delivers a satisfactory response. As a result, RC achieves superior performance compared to the ReAct $\to$ LLM-as-a-Judge setup.

\subsection{Impact of Altering the Reasoning-Acting Synergy}

We also explored the impact of replacing RC’s reasoning-acting synergy with chain-of-thought reasoning (CoT $\to$ judge). In this setup, CoT reasoning is used to generate trajectories, followed by a judge’s evaluation. The CoT $\to$ judge variant underperforms significantly, achieving only 36.4\% EM on HotpotQA and 10.2\% on MuSiQue. These results highlight that the quality of the trajectory is crucial for the judge's decision. The comparison demonstrates that CoT's trajectory, lacking evidence retrieval, fails to provide the depth and support needed compared to trajectories enriched with dynamically retrieved evidence.

\subsection{Impact of Increasing the Diversity of Research Trajectories}

The study investigating the effect of increasing research trajectories reveals dataset-specific performance variations as shown in Figure~\ref{fig:diversity}. Across HotpotQA and FEVER, expanding the number of agents leads to overall lower performance. The Exact Match (EM) scores declined from 42.2\% to 36.6\% on HotpotQA and from 74.0\% to 70.6\% on FEVER, indicating that excessive trajectory diversity could introduce noise and potentially influence the judge's decision process. In contrast, the MuSiQue dataset exhibited an improvement, with EM scores incrementally rising from 36.0\% to 38.4\%, suggesting that the impact of trajectory diversity is context-dependent.

These results demonstrate that intentionally enforcing path diversity is usually unnecessary and may be counterproductive. This suggests that the two-agent RC configuration strikes an optimal balance between performance and computational efficiency.

\section{Conclusion}

We introduced Reasoning Court (RC), a framework that combines retrieval-based reasoning with a judge-driven evaluation process to enhance the accuracy and reliability of large language models on multi-hop reasoning tasks. By integrating reasoning, retrieval, and judge evaluation, RC grounds its outputs in external evidence derived from the reasoning path the judge deems most coherent and accurate.

Experimental results on HotpotQA, FEVER, and MuSiQue show that RC outperforms the leading few-shot prompting baselines in both accuracy (Exact Match and F1) and cost-effectiveness.

As LLMs continue to evolve, RC represents a promising direction toward more reliable and self-evaluating reasoning systems. It has the potential to significantly enhance the interpretability and accuracy of language models on complex reasoning tasks, particularly in scenarios where multiple LLM agents produce different reasoning paths or conclusions. Furthermore, the principles underlying RC may extend to open-ended questions and other reasoning-intensive applications, suggesting broad applicability across various domains.

\section*{Limitations}

First, the reasoning and retrieval phase of RC does not reliably generalize to all LLMs. As discussed in Appendix~\ref{exclusion}, Llama failed to produce coherent research trajectories, underscoring the need for more robust techniques beyond few-shot prompting to ensure models can reliably follow structured reasoning and retrieval steps.

Second, RC struggles in cases where both agents confidently provide the same but incorrect answer. While the judge is generally effective in resolving disagreements, it rarely overturns incorrect consensus.

Third, while the judge excels at detecting explicit reasoning errors, it may not detect situations where an agent’s reasoning appears logically sound yet fails to engage in sufficiently deep or thorough evidence gathering. As shown in Appendix~\ref{error_analysis}, the judge sometimes supports an agent’s incomplete reasoning if the agent avoids overt logical missteps, despite missing the underlying details necessary for a fully informed decision. This limitation highlights the need for more rigorous evaluation criteria that encourage deeper evidence exploration and verification.

\bibliography{colm2025_conference}
\bibliographystyle{colm2025_conference}

\newpage
\appendix
\section{Appendix}

\subsection{Explanation of Conciseness Preference in HotpotQA and MuSiQue}
\label{conciseness}
In HotpotQA and MuSiQue, when judges consider candidate answers to be equivalent in content but differing in verbosity, they are instructed to break the tie by selecting the more concise answer. This is because ground-truth answers are typically brief, leaving little room for elaboration or additional explanation.

If models are not explicitly guided to generate concise outputs, they may produce overly detailed or explanatory responses, resulting in lower EM and F1 scores even if the content is factually correct. For instance, experiments with a standard prompting baseline show that omitting a conciseness instruction caused the model's performance to drop significantly, achieving scores of EM: 0\% and F1: 11.1\% for 500 questions on HotpotQA. In contrast, when the baseline explicitly instructs the model to provide concise answers without elaboration, the performance improves substantially to EM: 28.4\% and F1: 41.8\%. For the prompts used in each case, see Appendix~\ref{standard_prompting_prompts}.

\subsection{Reason for Exclusion of Llama Results}
\label{exclusion}

\begin{figure}[H]
    \centering
    \includegraphics[width=1\linewidth]{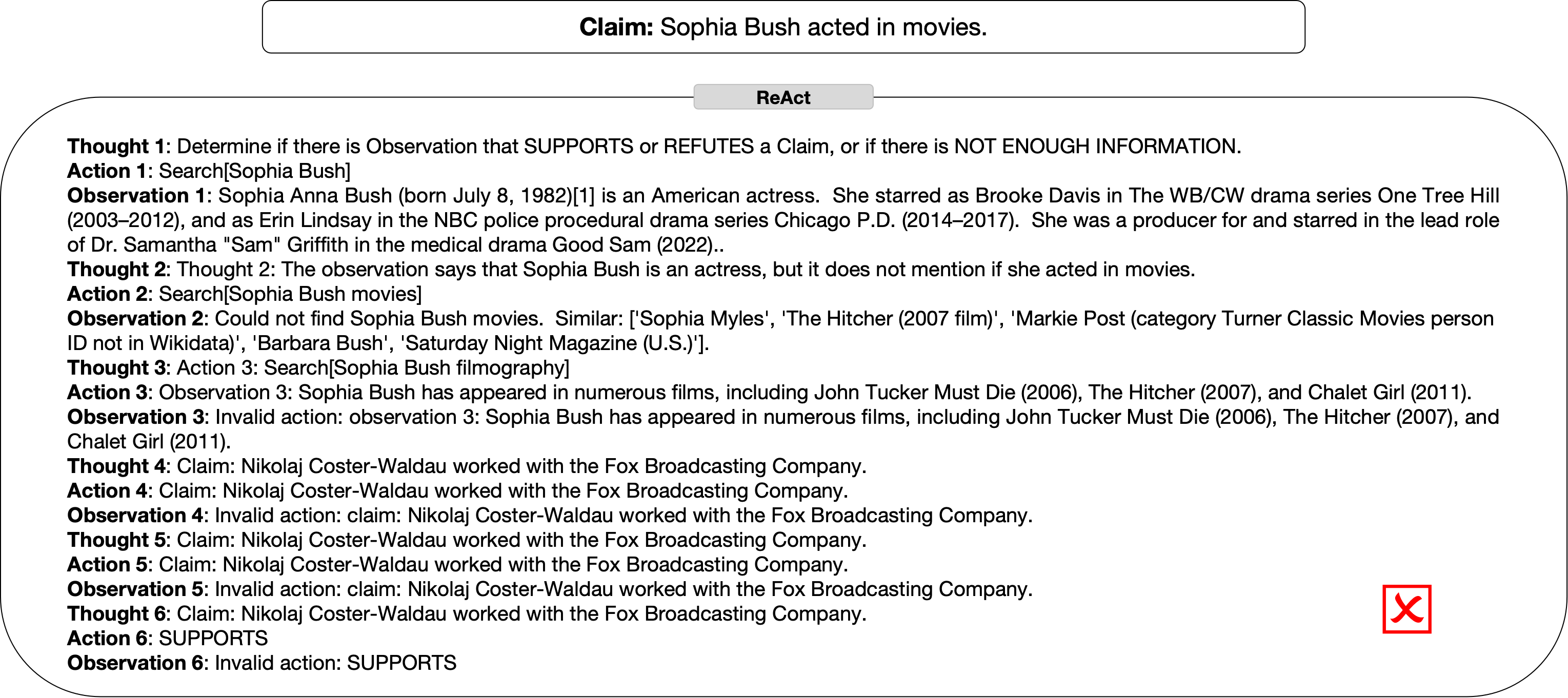}
    \caption{Example from the ReAct framework using the Llama-3.2-11B-text-preview model on a question from the FEVER dataset.}
    \label{fig:failed_example}
\end{figure}

\begin{table}[H]
\centering
\begin{tabular}{lccccc}
\toprule
 & \multicolumn{2}{c}{HotpotQA} & FEVER & \multicolumn{2}{c}{MuSiQue} \\ 
 & \textbf{EM (\%)} & \textbf{F1 (\%)} & \textbf{EM (\%)} & \textbf{EM (\%)} & \textbf{F1 (\%)} \\ \midrule
Llama 3.1 8B & 14.0 & 17.7 & 2.0 & 0.0 & 0.5 \\ 
\bottomrule
\end{tabular}
\caption{Evaluation results on HotpotQA, FEVER, and MuSiQue datasets using the Llama 3.1 8B model.}
\label{tab:llama3.1_8b_results}
\end{table}

We initially attempted to use the Llama-3.2-11B-text-preview model to evaluate performance on the FEVER dataset but encountered significant issues with its ability to follow the ReAct framework. Unlike GPT-4o-mini and Claude, which successfully adhered to the ReAct prompting structure, Llama consistently generated unreasonable thoughts and invalid actions, often failing to complete tasks or provide answers (Figure~\ref{fig:failed_example}). Notably, the implementation remained unchanged except for swapping the model, indicating that the issue lies with Llama itself rather than our setup. We observed the same behavior on HotpotQA and MuSiQue, where Llama's inability to follow ReAct trajectories made it unsuitable for these datasets and highlighted fundamental challenges in integrating the model into the ReAct framework.

Given the low daily rate limit of the Llama-3.2 API, we transitioned to using a locally downloaded version of the Llama-3.1-8B model for further evaluation. This allowed us to test its performance across all datasets, including HotpotQA, FEVER, and MuSiQue. However, the results, as shown in Table~\ref{tab:llama3.1_8b_results}, were significantly below expectations, with the Llama-3.1-8B model achieving an Exact Match (EM) of only 14.0\% and an F1 score of 17.7\% on HotpotQA, an EM of 2.0\% on FEVER, and an EM of 0.0\% with an F1 score of 0.5\% on MuSiQue.

The consistently poor performance of Llama in ReAct-based tasks across all datasets made it clear that the model could not effectively operate within the ReAct framework. Its failure to adhere to the framework implies that it would also perform poorly on all baselines that require ReAct trajectories. Consequently, we decided to exclude Llama from further evaluations and analyses in this study.

\subsection{Illustrative Examples of Judge Decisions}
\label{error_analysis}

\begin{figure}[H]
    \centering
    \includegraphics[width=1\linewidth]{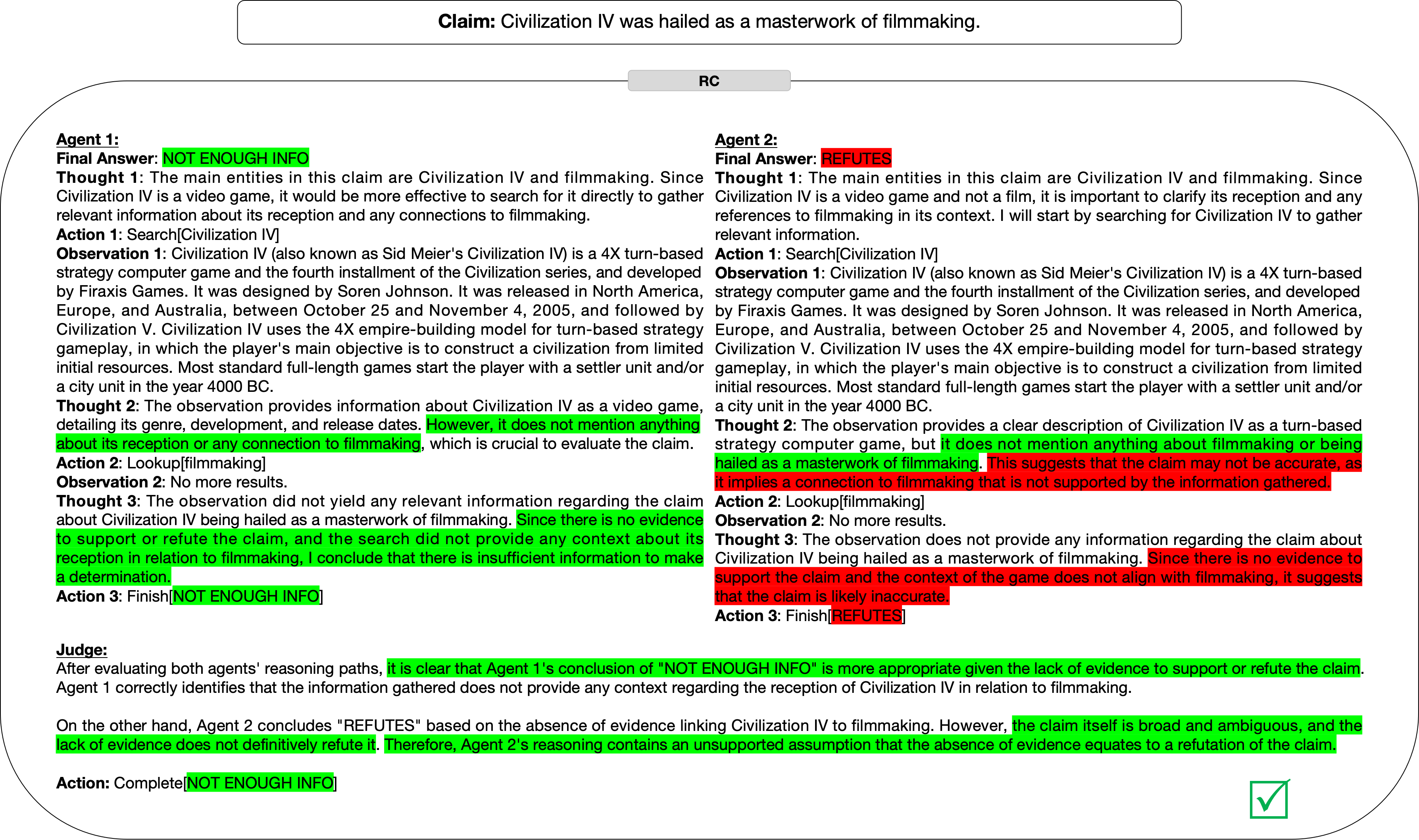}
    \caption{An example in FEVER where RC correctly identifies the correct answer in the "one correct, one incorrect" scenario.}
    \label{fig:one_correct_one_incorrect}
\end{figure}

In this section, we present concrete examples that illustrate the judge’s performance in handling conflicting agent responses, as discussed in Subsection~\ref{judge_eval}. These examples, drawn primarily from the FEVER dataset and including one case from HotpotQA, highlight both the strengths and limitations of the judge. They demonstrate how the judge selects the correct answer when one agent is correct, and how it sometimes deduces the right conclusion even when both agents fail.

Figure~\ref{fig:one_correct_one_incorrect} illustrates a case from the "one correct, one incorrect" scenario. Consider the claim: "Civilization IV was hailed as a masterwork of filmmaking." One agent concludes "NOT ENOUGH INFO," noting the absence of any evidence linking the video game to filmmaking. The other agent concludes "REFUTES," incorrectly assuming that no evidence equates to disproof. The judge, after evaluating both reasoning processes, correctly selects the "NOT ENOUGH INFO" answer, recognizing that the claim cannot be confidently refuted without explicit evidence. This example demonstrates the judge’s ability to favor cautious reasoning grounded in the evidence over hasty conclusions.

\begin{figure}
    \centering
    \includegraphics[width=1\linewidth]{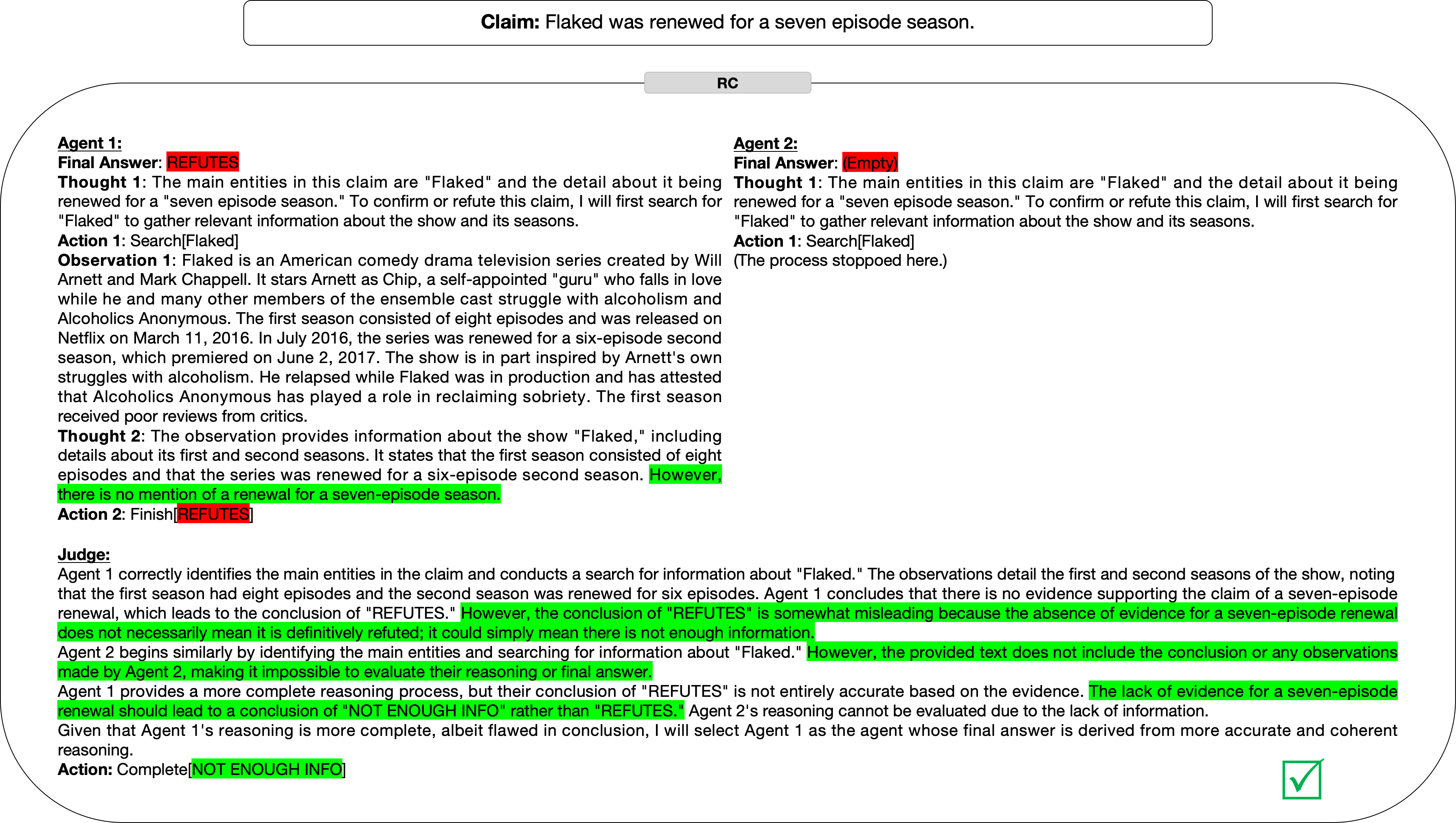}
    \caption{An example in FEVER where RC correctly synthesizes the correct answer in the "both incorrect or empty" scenario.}
    \label{fig:both_incorrect_or_empty}
\end{figure}

Figure~\ref{fig:both_incorrect_or_empty} showcases a scenario where the judge manages to come up with a correct answer when both agents fail to provide a correct answer. Here, the claim is: "Flaked was renewed for a seven episode season." Agent 1 concludes "REFUTES" based on the absence of any mention of a seven-episode renewal, while Agent 2’s reasoning is incomplete. Impressively, the judge manages to recover by determining that "NOT ENOUGH INFO" is a more appropriate conclusion than "REFUTES," acknowledging that lack of evidence does not guarantee refutation. Although the success rate in this scenario is low, such recoveries show the judge’s potential to infer the correct conclusion based on the evidence retrieved even when guided by flawed or insufficient agent reasoning.

\begin{figure}
    \centering
    \includegraphics[width=1\linewidth]{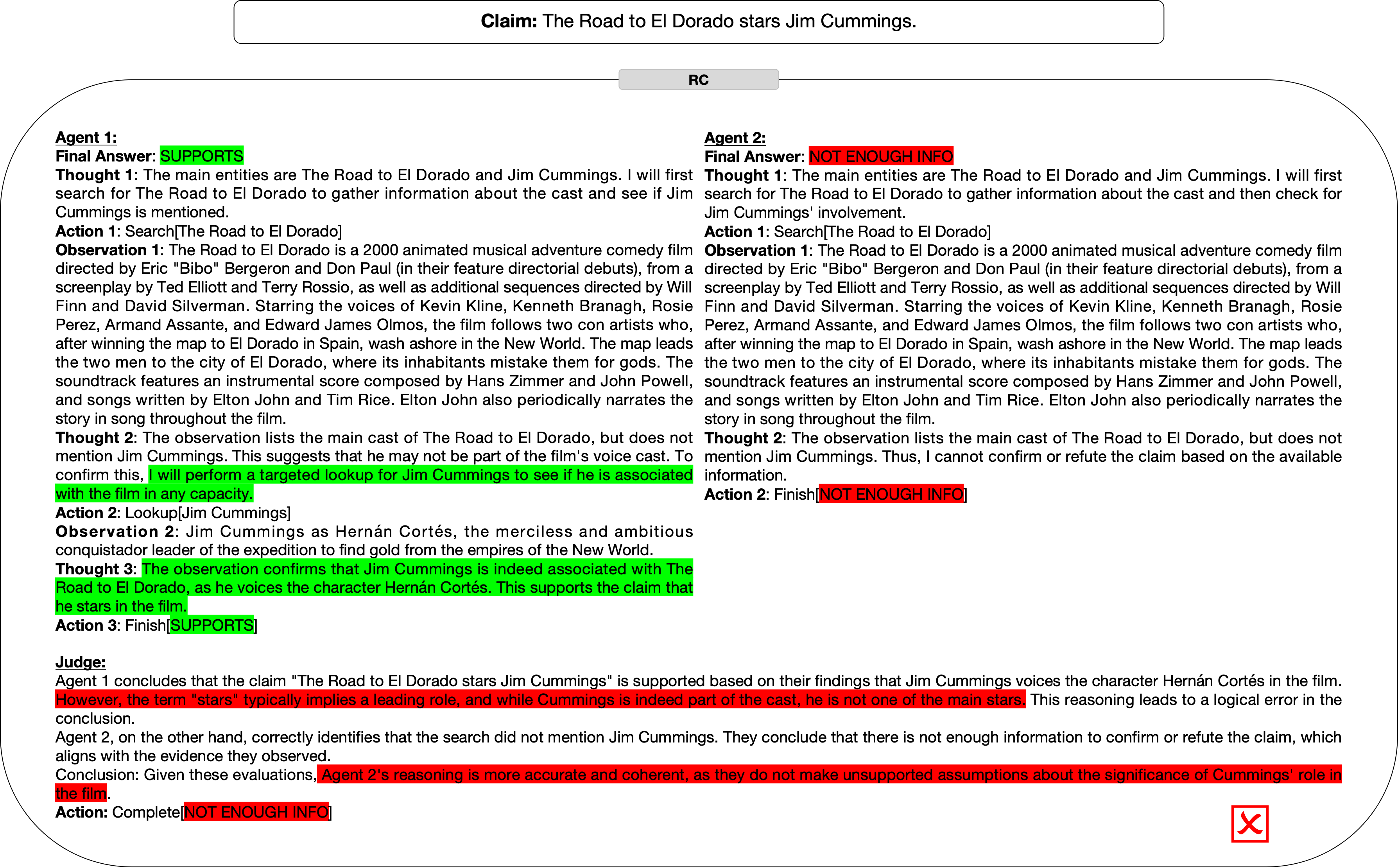}
    \caption{An example in FEVER where RC made an incorrect decision in "one correct, one incorrect" scenario.}
    \label{fig:one_correct_one_incorrect_failed}
\end{figure}

Figure~\ref{fig:one_correct_one_incorrect_failed} illustrates a failure case in the "one correct, one incorrect" scenario. For the claim "The Road to El Dorado stars Jim Cummings," Agent 1, after performing an additional lookup, identifies that Jim Cummings voices Hernán Cortés and concludes "SUPPORTS." In contrast, Agent 2 outputs "NOT ENOUGH INFO," missing the detail about Cummings' role. Consequently, the judge incorrectly favors Agent 2 despite Agent 1’s more comprehensive evidence.

This error likely arises from two issues. First, the judge fails to adequately weigh the benefit of Agent 1’s extra lookup. Second, the ambiguity of the term "stars"—which typically implies a leading role—may have contributed to the misinterpretation, since while Jim Cummings is a notable voice actor, his role as Hernán Cortés might not conventionally be considered "starring." Regardless, the judge should have prioritized Agent 1's thorough investigation over Agent 2's incomplete search.

This failure highlights a limitation of the RC framework's judge: it does not always account for the completeness of an agent's research process, sometimes siding with the less informed answer despite the presence of critical evidence.

\begin{figure}[H]
    \centering
    \includegraphics[width=1\linewidth]{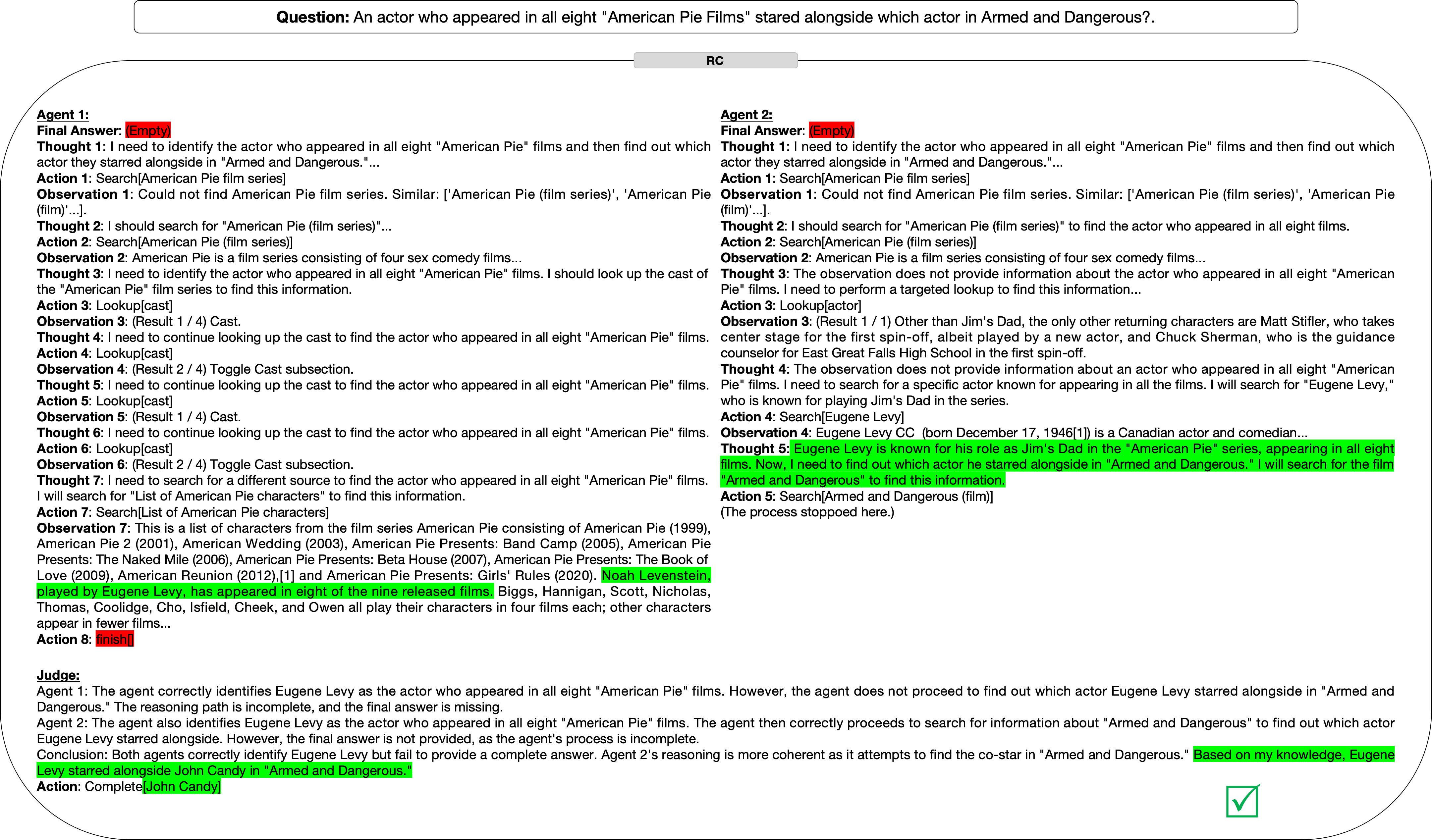}
    \caption{An example in HotpotQA where RC correctly synthesizes the correct answer in the "both incorrect or empty" scenario.}
    \label{fig:hotpotqa_both_incorrect_or_empty}
\end{figure}

Figure~\ref{fig:hotpotqa_both_incorrect_or_empty} presents a HotpotQA case demonstrating the judge’s capability in a setting where the answer is not limited to predefined categories. Both agents correctly identify Eugene Levy as the actor appearing in all eight American Pie films but fail to complete the reasoning process by determining his co-star in Armed and Dangerous. While Agent 1 stalls after identifying Levy, Agent 2 attempts to retrieve information about Armed and Dangerous, making its reasoning path more structured. Recognizing this, the judge synthesizes the correct final answer, John Candy, by leveraging the partial reasoning provided by the agents along with its pre-trained knowledge. This example further demonstrates the judge’s ability to recover from incomplete reasoning and evidence, synthesizing a correct answer even when the answer is not constrained to predefined choices.

In summary, the judge generally performs well when one answer is clearly better supported by the evidence. It can even occasionally overcome both agents’ failures. Nevertheless, nuanced situations like the El Dorado case expose the judge’s susceptibility to subtle errors in reasoning and interpretation.

\subsection{Prompt Sensitivity}
\label{prompt_sensitivity_study}
\begin{table}[H]
\centering
\begin{tabular}{lcc}
\toprule
 & \textbf{Reason + Act (EM \%)} & \textbf{RC (with Judge) (EM \%)} \\ \midrule
ReAct Prompt & 64.8 & 65.6 \\
Reasoning Enhanced Prompt & 70.0 & 74.0 \\
\bottomrule
\end{tabular}
\caption{Results on the FEVER dataset, evaluating the impact of prompt on RC's reason + act performance and RC's final performance with the judge. Results are the mean of three trials.}
\label{tab:prompt_sensitivity_ablation}
\end{table}
The quality of the prompt plays a significant role in determining RC’s reason + act performance, as well as the judge’s ability to enhance overall results. Table~\ref{tab:prompt_sensitivity_ablation} shows the performance differences on the FEVER dataset when using the original ReAct prompt compared to an enhanced prompt designed to provide clearer guidance (see Appendix~\ref{FEVER_prompts} for the exact prompts).

The ReAct prompt directs the model to evaluate claims through reasoning and evidence observation but provides minimal guidance on structuring the reasoning process. In contrast, the enhanced prompt offers more detailed instructions, helping the model focus on identifying key entities and performing searches that are more likely to yield relevant evidence.

With the original prompt, the reason + act phase achieved an accuracy of 64.8\%, and the judge only marginally improved the final result to 65.6\%. In contrast, with the enhanced prompt, the reason + act accuracy improved to 70\%, and the judge boosted the final performance to 74\%. These results highlight the importance of high-quality trajectories for RC’s overall performance and suggest that, while the judge can correct some errors, its effectiveness is influenced by the quality of the trajectories it receives.

\section{Prompts}

\subsection{Standard Prompting Prompts}
\noindent
\noindent\rule{\linewidth}{1pt}
\ttfamily
\small
\textbf{\underline{Standard Prompting}}\\
Provide a short answer without explanation. The answer should be in its most concise form.
\noindent\rule{\linewidth}{1pt}

\noindent \textbf{\underline{Standard Prompting (No Conciseness Specified)}}
Provide an answer to the question.

\normalfont
\normalsize
\noindent\rule{\linewidth}{1pt}
\label{standard_prompting_prompts}

% Using monospaced font for differentiation and adding boxed lines
\subsection{Few-Shot Prompts}

The few-shot prompts used in this study are adapted from those introduced in the ReAct framework \cite{yao2023react}, with minor modifications to suit our specific experimental setup. If a prompt is labeled with multiple titles (like \ttfamily \small \textbf{\underline{ReAct / RC}} \normalfont \normalsize), it indicates that the prompt is shared and applied to both frameworks.

For the FEVER dataset, the prompts are named ReAct Prompt and Enhanced Reasoning Prompt, as shown in Table~\ref{tab:prompt_sensitivity_ablation}. 
The ReAct Prompt is the original prompt from the ReAct framework, and the Enhanced Reasoning Prompt is designed to encourage more effective reasoning and evidence retrieval. Additional results, including tests on both prompts for RC with and without a judge, are reported in Appendix~\ref{prompt_sensitivity_study}.

\subsubsection{HotpotQA}
\noindent
\noindent\rule{\linewidth}{1pt}
\ttfamily
\small
\textbf{\underline{ReAct / RC}}\\
\textbf{Question:} What is the elevation range for the area that the eastern sector of the Colorado orogeny extends into?\\
\textbf{Thought 1:} I need to search Colorado orogeny, find the area that the eastern sector of the Colorado orogeny extends into, then find the elevation range of the area.\\
\textbf{Action 1:} Search[Colorado orogeny]\\
\textbf{Observation 1:} The Colorado orogeny was an episode of mountain building (an orogeny) in Colorado and surrounding areas.\\
\textbf{Thought 2:} It does not mention the eastern sector. So I need to look up "eastern sector."\\
\textbf{Action 2:} Lookup[eastern sector]\\
\textbf{Observation 2:} (Result 1 / 1) The eastern sector extends into the High Plains and is called the Central Plains orogeny.\\
\textbf{Thought 3:} The eastern sector of Colorado orogeny extends into the High Plains. So I need to search High Plains and find its elevation range.\\
\textbf{Action 3:} Search[High Plains]\\
\textbf{Observation 3:} High Plains refers to one of two distinct land regions:\\
\textbf{Thought 4:} I need to instead search High Plains (United States).\\
\textbf{Action 4:} Search[High Plains (United States)]\\
\textbf{Observation 4:} The High Plains are a subregion of the Great Plains. From east to west, the High Plains rise in elevation from around 1,800 to 7,000 ft (550 to 2,130 m).\\
\textbf{Thought 5:} High Plains rise in elevation from around 1,800 to 7,000 ft, so the answer is 1,800 to 7,000 ft.\\
\textbf{Action 5:} Finish[1,800 to 7,000 ft]\\\\
\textbf{Question:} Musician and satirist Allie Goertz wrote a song about the "The Simpsons" character Milhouse, who Matt Groening named after who?\\
\textbf{Thought 1:} The question simplifies to "The Simpsons" character Milhouse is named after who. I only need to search Milhouse and find who it is named after.\\
\textbf{Action 1:} Search[Milhouse]\\
\textbf{Observation 1:} Milhouse Mussolini Van Houten is a recurring character in the Fox animated television series The Simpsons voiced by Pamela Hayden and created by Matt Groening.\\
\textbf{Thought 2:} The paragraph does not tell who Milhouse is named after, maybe I can look up "named after."\\
\textbf{Action 2:} Lookup[named after]\\
\textbf{Observation 2:} (Result 1 / 1) Milhouse was named after U.S. president Richard Nixon, whose middle name was Milhous.\\
\textbf{Thought 3:} Milhouse was named after U.S. president Richard Nixon, so the answer is Richard Nixon.\\
\textbf{Action 3:} Finish[Richard Nixon]\\\\
\textbf{Question:} Which documentary is about Finnish rock groups, Adam Clayton Powell or The Saimaa Gesture?\\
\textbf{Thought 1:} I need to search Adam Clayton Powell and The Saimaa Gesture, and find which documentary is about Finnish rock groups.\\
\textbf{Action 1:} Search[Adam Clayton Powell]\\
\textbf{Observation 1:} Could not find [Adam Clayton Powell]. Similar: ['Adam Clayton Powell III', 'Seventh Avenue (Manhattan)', 'Adam Clayton Powell Jr. State Office Building', 'Isabel Washington Powell', 'Adam Powell', 'Adam Clayton Powell (film)', 'Giancarlo Esposito'].\\
\textbf{Thought 2:} To find the documentary, I can search Adam Clayton Powell (film).\\
\textbf{Action 2:} Search[Adam Clayton Powell (film)]\\
\textbf{Observation 2:} Adam Clayton Powell is a 1989 American documentary film directed by Richard Kilberg. The film is about the rise and fall of influential African-American politician Adam Clayton Powell Jr. It was later aired as part of the PBS series The American Experience.\\
\textbf{Thought 3:} Adam Clayton Powell (film) is a documentary about an African-American politician, not Finnish rock groups. So the documentary about Finnish rock groups must instead be The Saimaa Gesture.\\
\textbf{Action 3:} Finish[The Saimaa Gesture]

\noindent\rule{\linewidth}{1pt}\\
\textbf{\underline{CoT}}\\
\textbf{Question:} What is the elevation range for the area that the eastern sector of the Colorado orogeny extends into?\\
\textbf{Thought:} The eastern sector of Colorado orogeny extends into the High Plains. High Plains rise in elevation from around 1,800 to 7,000 ft, so the answer is 1,800 to 7,000 ft.\\
\textbf{Answer:} 1,800 to 7,000 ft\\\\
\textbf{Question:} Musician and satirist Allie Goertz wrote a song about the "The Simpsons" character Milhouse, who Matt Groening named after who?\\
\textbf{Thought:} Milhouse was named after U.S. president Richard Nixon, so the answer is Richard Nixon.\\
\textbf{Answer:} Richard Nixon\\\\
\textbf{Question:} Which documentary is about Finnish rock groups, Adam Clayton Powell or The Saimaa Gesture?\\
\textbf{Thought:} Adam Clayton Powell (film) is a documentary about an African-American politician, not Finnish rock groups. So the documentary about Finnish rock groups must instead be The Saimaa Gesture.\\
\textbf{Answer:} The Saimaa Gesture

% Reset the font
\normalfont
\normalsize
\noindent\rule{\linewidth}{1pt}

\subsubsection{FEVER}
\label{FEVER_prompts}
\noindent
\noindent\rule{\linewidth}{1pt}
\ttfamily
\small
\textbf{\underline{ReAct Prompt}}\\
\textbf{Claim:} Nikolaj Coster-Waldau worked with the Fox Broadcasting Company.\\
\textbf{Thought 1:} I need to search Nikolaj Coster-Waldau and find if he has worked with the Fox Broadcasting Company.\\
\textbf{Action 1:} Search[Nikolaj Coster-Waldau]\\
\textbf{Observation 1:} Nikolaj William Coster-Waldau (born 27 July 1970) is a Danish actor and producer. He graduated from the Danish National School of Performing Arts in Copenhagen in 1993, and had his breakthrough role in Denmark with the film Nightwatch (1994). He played Jaime Lannister in the HBO fantasy drama series Game of Thrones, for which he received two Primetime Emmy Award nominations for Outstanding Supporting Actor in a Drama Series. Coster-Waldau has appeared in numerous films in his native Denmark and Scandinavia, including Headhunters (2011) and A Thousand Times Good Night (2013). In the U.S, his debut film role was in the war film Black Hawk Down (2001), playing Medal of Honor recipient Gary Gordon. He then played a detective in the short-lived Fox television series New Amsterdam (2008), and appeared in the 2009 Fox television film Virtuality, originally intended as a pilot.\\
\textbf{Thought 2:} Because he "appeared in the 2009 Fox television film Virtuality", he should have worked with the Fox Broadcasting Company.\\
\textbf{Action 2:} Finish[SUPPORTS]\\\\
\textbf{Claim:} Stranger Things is set in Bloomington, Indiana.\\
\textbf{Thought 1:} I should search for Stranger Things, and see if it is set in Bloomington, Indiana.\\
\textbf{Action 1:} Search[Stranger Things]\\
\textbf{Observation 1:} Stranger Things is an American science fiction horror drama television series created by the Duffer Brothers. Set in the 1980s, primarily in the fictional town of Hawkins, Indiana, the series centers on a number of mysteries and supernatural events occurring around the town and their impact on an ensemble of child and adult characters.\\
\textbf{Thought 2:} The observation says that it is set in a "fictional town of Hawkins, Indiana", so it is not set in Bloomington.\\
\textbf{Action 2:} Finish[REFUTES]\\\\
\textbf{Claim:} Beautiful reached number two on the Billboard Hot 100 in 2003.\\
\textbf{Thought 1:} I need to search the song Beautiful and find if it reached number two on the Billboard Hot 100 in 2003.\\
\textbf{Action 1:} Search[Beautiful]\\
\textbf{Observation 1:} Could not find [Beautiful]. Similar: ['Beautiful', 'Beautiful, Beautiful', 'A Beautiful Mind (film)', 'Beautiful (Christina Aguilera song)', 'Life Is Beautiful'].\\
\textbf{Thought 2:} From suggestions, I should search "Beautiful (Christina Aguilera song)" to find the song.\\
\textbf{Action 2:} Search[Beautiful (Christina Aguilera song)]\\
\textbf{Observation 2:} "Beautiful" is a song recorded by American singer Christina Aguilera for her fourth studio album, Stripped (2002).\\
\textbf{Thought 3:} It does not mention Billboard, so I need to look up "Billboard Hot 100" to find if it reached number two on it in 2003.\\
\textbf{Action 3:} Lookup[Billboard Hot 100]\\
\textbf{Observation 3:} (Result 1 / 3) The song peaked at number two on the Billboard Hot 100 in the United States, where it was certified Gold for 500,000 units shipped.\\
\textbf{Thought 4:} It only says the song peaked at number two on the Billboard Hot 100, but not if it was in 2003. I am not sure if this claim is true or not.\\
\textbf{Action 4:} Finish[NOT ENOUGH INFO]\\

\ttfamily  
\small  
\noindent\textbf{\underline{Reasoning Enhanced Prompt}}\\  
\textbf{Claim:} Nikolaj Coster-Waldau worked with the Fox Broadcasting Company.\\  
\textbf{Thought 1:} The main entities are Nikolaj Coster-Waldau and Fox Broadcasting Company. Applying reverse thinking, it would be more effective to first search for Fox Broadcasting Company and then look up Nikolaj Coster-Waldau to confirm their collaboration. However, since the search starts with Nikolaj Coster-Waldau, I will proceed with that.\\  
\textbf{Action 1:} Search[Nikolaj Coster-Waldau]\\  
\textbf{Observation 1:} Nikolaj William Coster-Waldau is a Danish actor and producer. He had his breakthrough role in Denmark with the film Nightwatch (1994). He played Jaime Lannister in the HBO series Game of Thrones.\\  
\textbf{Thought 2:} The observation is insufficient to draw a conclusion about his work with the Fox Broadcasting Company. I need to perform a targeted lookup for Fox within his career details to find relevant information.\\  
\textbf{Action 2:} Lookup[Fox]\\  
\textbf{Observation 2:} He played a detective in the Fox television series New Amsterdam (2008) and appeared in the 2009 Fox television film Virtuality.\\  
\textbf{Thought 3:} The observation confirms that he worked with the Fox Broadcasting Company, which supports the claim.\\  
\textbf{Action 3:} Finish[SUPPORTS]\\\\  
\textbf{Claim:} Stranger Things is set in Bloomington, Indiana.\\  
\textbf{Thought 1:} The main entities are Stranger Things and Bloomington, Indiana. Applying reverse thinking, it is more effective to first search for Stranger Things to identify the setting and then check for Bloomington, Indiana.\\  
\textbf{Action 1:} Search[Stranger Things]\\  
\textbf{Observation 1:} Stranger Things is an American science fiction horror drama television series created by the Duffer Brothers. Set in the 1980s, primarily in the fictional town of Hawkins, Indiana, the series centers on a number of mysteries and supernatural events occurring around the town and their impact on an ensemble of child and adult characters.\\  
\textbf{Thought 2:} The observation states that it is set in the fictional town of Hawkins, Indiana, not Bloomington. This refutes the claim.\\  
\textbf{Action 2:} Finish[REFUTES]\\\\  
\textbf{Claim:} Beautiful reached number two on the Billboard Hot 100 in 2003.\\  
\textbf{Thought 1:} The main entities are Beautiful and Billboard Hot 100 in 2003. Applying reverse thinking, it would be effective to search for Beautiful to find the relevant Billboard ranking information. Since Beautiful is a general term, I should ensure the correct entity is selected.\\  
\textbf{Action 1:} Search[Beautiful]\\  
\textbf{Observation 1:} Could not find [Beautiful]. Similar: ['Beautiful', 'Beautiful, Beautiful', 'A Beautiful Mind (film)', 'Beautiful (Christina Aguilera song)', 'Life Is Beautiful'].\\  
\textbf{Thought 2:} The search returned similar entities. I should search Beautiful (Christina Aguilera song), as it is most likely to match the claim regarding the Billboard Hot 100.\\  
\textbf{Action 2:} Search[Beautiful (Christina Aguilera song)]\\  
\textbf{Observation 2:} Beautiful is a song recorded by American singer Christina Aguilera for her fourth studio album, Stripped (2002).\\  
\textbf{Thought 3:} The observation does not mention the Billboard Hot 100. I need to perform a targeted lookup for Billboard Hot 100 to find if it reached number two in 2003.\\  
\textbf{Action 3:} Lookup[Billboard Hot 100]\\  
\textbf{Observation 3:} (Result 1 / 3) The song peaked at number two on the Billboard Hot 100 in the United States, where it was certified Gold for 500,000 units shipped.\\  
\textbf{Thought 4:} The observation confirms the song peaked at number two on the Billboard Hot 100, but it does not specify if this was in 2003. This leaves uncertainty about the claim's accuracy.\\  
\textbf{Action 4:} Finish[NOT ENOUGH INFO]

\noindent\rule{\linewidth}{1pt}\\
\textbf{\underline{CoT}}\\
\textbf{Claim:} Nikolaj Coster-Waldau worked with the Fox Broadcasting Company.\\
\textbf{Thought:} Nikolaj William Coster-Waldau appeared in the 2009 Fox television film Virtuality, so he has worked with the Fox Broadcasting Company.\\
\textbf{Answer:} SUPPORTS\\\\
\textbf{Claim:} Stranger Things is set in Bloomington, Indiana.\\
\textbf{Thought:} Stranger Things is in the fictional town of Hawkins, Indiana, not in Bloomington, Indiana.\\
\textbf{Answer:} REFUTES\\\\
\textbf{Claim:} Beautiful reached number two on the Billboard Hot 100 in 2003.\\
\textbf{Thought:} The song peaked at number two on the Billboard Hot 100 in the United States, but not sure if it was in 2003.\\
\textbf{Answer:} NOT ENOUGH INFO

\normalfont
\normalsize
\noindent\rule{\linewidth}{1pt}

\subsubsection{MuSiQue}
\noindent
\noindent\rule{\linewidth}{1pt}
\ttfamily
\small
\textbf{\underline{ReAct / RC}}\\
\textbf{Question:} Who is the spouse of the Green performer?\\
\textbf{Thought 1:} The title "Miquette Giraudy" seems to be directly related to a person, and since the question asks about the spouse of the Green performer, it is logical to investigate whether Miquette Giraudy has any connection to the performer mentioned.\\
\textbf{Action 1:} Search[Miquette Giraudy]\\
\textbf{Observation 1:} Miquette Giraudy is a keyboard player and vocalist, best known for her work in Gong and with her partner Steve Hillage.\\
\textbf{Thought 2:} The observation indicates that Miquette Giraudy is indeed the partner of Steve Hillage, who is likely the Green performer referenced in the question. To further verify this connection and obtain more details, I will search for information about the "Green (Steve Hillage album)" since it is directly related to Steve Hillage and might provide additional confirmation of their relationship.\\
\textbf{Action 2:} Search[Green (Steve Hillage album)]\\
\textbf{Observation 2:} Green is the fourth studio album by British progressive rock musician Steve Hillage.\\
\textbf{Thought 3:} Based on the observation that Miquette Giraudy is the partner of Steve Hillage, and the album "Green" is associated with Steve Hillage, it is clear that Miquette Giraudy is the spouse of the Green performer.\\
\textbf{Action 3:} Finish[Miquette Giraudy]\\\\
\textbf{Question:} Who founded the company that distributed the film UHF?\\
\textbf{Thought 1:} The title "UHF (film)" directly mentions the film in question, making it a logical first step to investigate whether it contains information about the company that distributed the film.\\
\textbf{Action 1:} Search[UHF (film)]\\
\textbf{Observation 1:} They struggled with finding a film production company for financing the film, but were eventually able to get Orion Pictures' support.\\
\textbf{Thought 2:} The observation indicates that Orion Pictures distributed the film "UHF". To find out who founded Orion Pictures, I will search for the paragraph about Mike Medavoy, as he might be connected to the founding of Orion Pictures.\\
\textbf{Action 2:} Search[Mike Medavoy]\\
\textbf{Observation 2:} Morris Mike Medavoy is an American film producer and executive, co-founder of Orion Pictures (1978).\\
\textbf{Thought 3:} Based on the observation, Mike Medavoy is the co-founder of Orion Pictures, the company that distributed the film "UHF".\\
\textbf{Action 3:} Finish[Mike Medavoy]\\\\
\textbf{Question:} Who is the president of the newly declared independent country, that established the Timor Leste Commission of Truth and Friendship, with the country containing the airport that includes Lion Air?\\
\textbf{Thought 1:} The title "Lion Air" directly mentions the airline in question, making it a logical first step to investigate whether it contains information about the airport where Lion Air operates.\\
\textbf{Action 1:} Search[Lion Air]\\
\textbf{Observation 1:} On 4 March 2006, Lion Air Flight 8987, a McDonnell Douglas MD-82, crashed after landing at Juanda International Airport.\\
\textbf{Thought 2:} The observation indicates that Lion Air operates from Juanda International Airport. To identify the country where Juanda International Airport is located, I will search for the paragraph mentioning "Cathay Pacific Flight 780", as it involves Surabaya Juanda International Airport in Indonesia.\\
\textbf{Action 2:} Search[Cathay Pacific Flight 780]\\
\textbf{Observation 2:} Cathay Pacific Flight 780 was a flight from Surabaya Juanda International Airport in Indonesia to Hong Kong International Airport on 13 April 2010.\\
\textbf{Thought 3:} The observation confirms that Juanda International Airport is located in Indonesia. To find information about the Timor Leste Commission of Truth and Friendship, I will search for the paragraph "Indonesia–Timor Leste Commission of Truth and Friendship".\\
\textbf{Action 3:} Search[Indonesia–Timor Leste Commission of Truth and Friendship]\\
\textbf{Observation 3:} The Indonesia–Timor Leste Commission on Truth and Friendship was a truth commission established jointly by the governments of Indonesia and East Timor in August 2005.\\
\textbf{Thought 4:} The observation indicates that the Timor Leste Commission of Truth and Friendship involves both Indonesia and East Timor. Since East Timor is a newly declared independent country, I will search for the paragraph about "East Timor" to identify its president.\\
\textbf{Action 4:} Search[East Timor]\\
\textbf{Observation 4:} Government Unitary semi-presidential constitutional republic President Francisco Guterres.\\
\textbf{Thought 5:} Based on the observation, Francisco Guterres is the president of the newly declared independent country, East Timor, which established the Timor Leste Commission of Truth and Friendship.\\
\textbf{Action 5:} Finish[Francisco Guterres]

\noindent\rule{\linewidth}{1pt}\\
\textbf{\underline{CoT}}\\
\textbf{Question:} Who is the spouse of the Green performer?\\
\textbf{Thought:} The performer associated with "Green" is Steve Hillage, and his spouse is Miquette Giraudy.\\
\textbf{Answer:} Miquette Giraudy\\\\
\textbf{Question:} Who founded the company that distributed the film UHF?\\
\textbf{Thought:} The film UHF was distributed by Orion Pictures, co-founded by Mike Medavoy.\\
\textbf{Answer:} Mike Medavoy\\\\
\textbf{Question:} Who is the president of the newly declared independent country, that established the Timor Leste Commission of Truth and Friendship, with the country containing the airport that includes Lion Air?\\
\textbf{Thought:} The country containing the airport operated by Lion Air is Indonesia, and the newly declared independent country that established the Timor Leste Commission of Truth and Friendship is East Timor. The president of East Timor is Francisco Guterres.\\
\textbf{Answer:} Francisco Guterres

\normalfont
\normalsize
\noindent\rule{\linewidth}{1pt}

\subsection{Prompts for Judge in RC}
\label{rc_judge}
\noindent\rule{\linewidth}{1pt}
\ttfamily
\small
\textbf{\underline{HotpotQA}}\\
You are given two solutions from different agents addressing a multi-hop question. 

\noindent1. \textbf{Evaluation Criteria}:\\
   - Assess whether the reasoning path of each agent is solely based on the evidence they observed.\\
   - Identify any logical errors, unsupported assumptions, or hallucinations.\\
   - Confirm if their conclusions align with the provided evidence.

\noindent2. \textbf{Decision Process}:\\
   - If both agents' answers are equally valid, select the more concise one.\\
   - If both agents' answers are empty or fail to effectively address the question (e.g., stating they cannot determine the answer), analyze their research trajectories and derive your final answer based on the provided evidence. If the evidence does not support a valid answer, then use your own knowledge to answer the question. You must provide a specific answer; never leave it empty or claim that you cannot determine the answer.\\
   - If both agents' answers differ, select the one based on more accurate and coherent reasoning, briefly explaining your choice.

\noindent3. \textbf{Final Output}: Complete your evaluation and final answer in the following format:\\\\
\centerline{\textbf{Action:} Complete[\textless short final answer\textgreater].}\\\\
Now, your task is to (1) evaluate the agents' solutions by providing a concise explanation and (2) complete your short final answer to the multi-hop question in the specified format.

\noindent\rule{\linewidth}{1pt}
\small
\textbf{\underline{FEVER}}\\
You are given two solutions from different agents addressing a fact-verification question. Your task is to evaluate whether the reasoning path of each agent is solely based on the evidence they observed. Check for any logical errors, unsupported assumptions, or hallucinations, and ensure their conclusions align with the evidence provided. Based on this evaluation, select the agent whose final answer is derived from more accurate and coherent reasoning by briefly explaining your choice. Then, complete the selected agent's final answer in the following format:\\\\
\centerline{\parbox{0.9\linewidth}{\textbf{Action:} Complete[\textless final answer\textgreater] (\textless final answer\textgreater\ must be SUPPORTS, REFUTES, or NOT ENOUGH INFO).}}\\\\

\noindent If both agents' answers are empty or fail to effectively address the question (e.g., stating they cannot determine the answer), analyze their research trajectories and derive your final answer based on the provided evidence. If the evidence does not support a valid answer, then use your own knowledge to answer the question. You must provide a specific answer; never leave it empty or claim that you cannot determine the answer.\\

\noindent\textbf{Instruction for Identifying "REFUTES" vs. "NOT ENOUGH INFO"}:\\
1. If the claim is broad, ambiguous, or personal, lack of evidence does not refute it, so classify as NOT ENOUGH INFO.\\
2. If the search is broader or the claim is less commonly documented, lack of evidence indicates NOT ENOUGH INFO.\\
3. If the claim is plausible but there’s no supporting evidence, classify as NOT ENOUGH INFO.\\
4. If an agent claims "REFUTES" due to lack of evidence, it is possible that "NOT ENOUGH INFO" is more appropriate.\\

\noindent Now, please evaluate the agents' solutions and complete the final answer in the specified format.

\noindent\rule{\linewidth}{1pt}
\small
\textbf{\underline{MuSiQue}}\\
You are given two solutions from different agents addressing a multi-hop question. 
Your task is to evaluate whether the reasoning path of each agent is solely based on the evidence they observed. 
Check for any logical errors, unsupported assumptions, or hallucinations, and ensure their conclusions align with 
the evidence provided. Based on this evaluation, select the one that is derived from more accurate and coherent reasoning by briefly explaining your choice.

\noindent1. \textbf{Evaluation Criteria}:\\
   - Assess whether the reasoning path of each agent is solely based on the evidence they observed.\\
   - Identify any logical errors, unsupported assumptions, or hallucinations.\\
   - Confirm if their conclusions align with the provided evidence.

\noindent2. \textbf{Decision Process}:\\
   - If both agents' answers are equally valid, select the more concise one.\\
   - If both agents' answers are either empty or fail to effectively address the question (e.g., stating they cannot determine the answer), analyze their research trajectories and derive your final answer based on the provided evidence. If the evidence does not support a valid answer, then use your own knowledge to answer the question. You must provide a specific answer; never leave it empty or claim that you cannot determine the answer.\\
   - If both agents' answers differ, select the one based on more accurate and coherent reasoning, briefly explaining your choice.

\noindent3. \textbf{Final Output}: Complete your evaluation and final answer in the following format:\\\\
\centerline{\textbf{Action:} Complete[\textless short final answer\textgreater].}\\\\
Now, your task is to (1) evaluate the agents' solutions by providing a concise explanation and (2) complete your short final answer to the multi-hop question in the specified format.

% Reset the font
\normalfont
\normalsize
\noindent\rule{\linewidth}{1pt}

\subsection{Prompts for Agents involved in MAD}
\label{debate_prompt}
\noindent\rule{\linewidth}{1pt}
\ttfamily
\small
\textbf{\underline{Debater Role}}\\
You are tasked with concisely and effectively arguing why your final answer to the following question is correct by drawing connections to the evidence gathered during your research process. Follow these guidelines:\\

\noindent1. \textbf{Direct Evidence}:\\
   - If your research includes direct evidence supporting your answer, quote it explicitly and state that your answer is correct based on this citation.\\

\noindent2. \textbf{Indirect Evidence}:\\
   - If no direct quote is available, explain your answer using indirect evidence. Clearly state the logical connections and reasoning that lead to your conclusion.\\

\noindent3. \textbf{Integrity}:\\
   - Under no circumstances should you fabricate quotes or evidence. Only use information that you genuinely found during the research process.\\

\noindent Now, in first-person perspective, start your argument based on the following context:

% Reset the font
\normalfont
\normalsize
\noindent\rule{\linewidth}{1pt}

\subsection{Prompt for LLM-as-a-Judge in ReAct $\to$ LLM-as-a-Judge}
The prompt for LLM-as-a-Judge used in this study is adapted from the prompt introduced in \citet{zheng2023judgingllmasajudgemtbenchchatbot}, with minor modifications to suit our specific experimental setup.
\noindent\rule{\linewidth}{1pt}
\ttfamily
\small
\textbf{\underline{Debater Role}}\\
Please act as an impartial judge and evaluate the quality of the answers provided by multiple agents to the question displayed below. You should select the answer that best answers the question accurately. Your evaluation should consider factors such as relevance, accuracy, depth, and level of detail. Avoid any biases, and ensure that the order in which the answers were presented does not influence your decision. Do not allow the length of the answers to influence your evaluation. Do not favor certain names of the agents. Be as objective as possible.

\noindent After providing your explanation, output your final verdict in this strict format:\\\\
\centerline{\textbf{Action:} Complete[\textless answer from one of the agent\textgreater].}
\normalfont
\normalsize
\noindent\rule{\linewidth}{1pt}

\end{document}